\documentclass{article}

\PassOptionsToPackage{compress}{natbib}
\usepackage[nonatbib,final]{neurips_2021}

\usepackage[utf8]{inputenc} 
\usepackage[T1]{fontenc}    
\usepackage[colorlinks]{hyperref}       
\usepackage{url}            
\usepackage{booktabs}       
\usepackage{amsfonts}       
\usepackage{nicefrac}       
\usepackage{microtype}      
\usepackage{subfiles}
\usepackage{graphicx}
\usepackage{wrapfig}

\usepackage{subfigure}
\usepackage{multirow}
\usepackage{amsmath}
\usepackage{mathtools}
\usepackage{xcolor,pifont}
\usepackage{cite}
\usepackage{float}
\usepackage{slashbox}
\usepackage{tikz}
\usetikzlibrary{calc}

\newcommand{\tabincell}[2]{\begin{tabular}{@{}#1@{}}#2\end{tabular}}
\newcommand\tikzmark[1]{%
  \tikz[overlay,remember picture] \coordinate (#1);}

\newcommand{\firstkey}[1]{\textcolor{red}{\textbf{#1}}}
\newcommand{\secondkey}[1]{\textcolor{blue}{\textbf{#1}}}

\newcommand{\Section}[1]{\vspace{-2mm} \section{#1} \vspace{-1mm}}
\newcommand{\SubSection}[1]{\vspace{-1mm} \subsection{#1} \vspace{-1mm}}

\newcommand{\Paragraph}[1]{\vspace{0mm} \noindent \textbf{#1} \hspace{0mm}}
\usepackage{tikz}
\usetikzlibrary{calc}
\newcommand*\circled[1]{\tikz[baseline=(char.base)]{
    \node[shape=circle, draw, inner sep=1pt, 
        minimum height={\f@size*1.4},] (char) {\vphantom{WAH1g}#1};}}
\newcommand{\greencheck}{\textcolor{black}{\ding{51}}} 

\title{Voxel-based 3D Detection and Reconstruction of Multiple Objects from a Single Image}

\author{%
  Feng Liu  $\enskip \! \! $ $\enskip \! \! $ $\enskip \! \! $  $\enskip \! \! $
  Xiaoming Liu \\
  
Department of Computer Science and Engineering \\
Michigan State University, East Lansing MI 48824\\
\texttt{\{liufeng6, liuxm\}@msu.edu}
}

\begin{document}

\maketitle


\begin{abstract}

Inferring $3$D locations and shapes of multiple objects from a single $2$D image is a long-standing objective of computer vision. 
Most of the existing works either predict one of these $3$D properties or focus on solving both for a single object. 
One fundamental challenge lies in how to learn an effective representation of the image that is well-suited for $3$D detection and reconstruction. 
In this work, we propose to learn a regular grid of $3$D voxel features from the input image which is aligned with $3$D scene space via a $3$D feature lifting operator. 
Based on the $3$D voxel features, our novel CenterNet-3D detection head formulates the $3$D detection as keypoint detection in the $3$D space. 
Moreover, we devise an efficient coarse-to-fine reconstruction module, including coarse-level voxelization and a novel local PCA-SDF shape representation, which enables fine detail reconstruction and one order of magnitude faster inference than prior methods. 
With complementary supervision from both $3$D detection and reconstruction, one enables the $3$D voxel features to be geometry and context preserving, benefiting both tasks.
The effectiveness of our approach is demonstrated through $3$D detection and reconstruction in single object and multiple object scenarios. 
Code is available at \url{http://cvlab.cse.msu.edu/project-mdr.html}.
\end{abstract}


\Section{Introduction}\label{sec:intro}
As a fundamental computer vision task, instance-level $3$D scene understanding from a single image has drawn substantial attention from researchers due to its importance in applications such as robotics~\cite{guerry2017snapnet,tateno2017cnn}, AR/VR~\cite{han2020live} and autonomous driving~\cite{behl2017bounding,chen2018lidar,geiger2012we}. 
The important $3$D properties include $3$D bounding box (pose, size, location) and $3$D shape of  object instances. 
In this work, we aim to design a framework to infer all these $3$D properties of multiple objects from a single $2$D image.  

In recent years, various monocular methods are proposed to predict either $3$D boxes~\cite{chen2016monocular,huang2018cooperative,brazil2019m3d,chen2020monopair,kinematic-3d-object-detection-in-monocular-video,groomed-nms-grouped-mathematically-differentiable-nms-for-monocular-3d-object-detection} or $3$D shapes~\cite{wu2017marrnet,zhu2018visual,chen2018learning,groueix2018atlasnet,wen2019pixel2mesh++}. However, only a few studies\cite{kundu20183d,nie2020total3dunderstanding,runz2020frodo,meshrcnn,popov2020corenet,engelmann2020points} consider both $3$D detection and reconstruction for a total $3$D scene understanding.
The complexity of real-world scenarios and diverse category variations make it challenging to fully reconstruct the scene context (both semantics and geometry) at the instance level from a single image.
Moreover, those methods primarily assign $3$D semantic labels to pixels. Yet, such a $2$D representation with depth ambiguity is insufficient for $3$D geometry and context reasoning. 
It is thus crucial to develop an effective representation of the image 
that is relevant to $3$D geometry and spatial information for performing accurate $3$D detection and reconstruction.

\begin{figure}[t!]
\centering
\includegraphics[trim=0 0 0 0,clip,width=13.8cm]{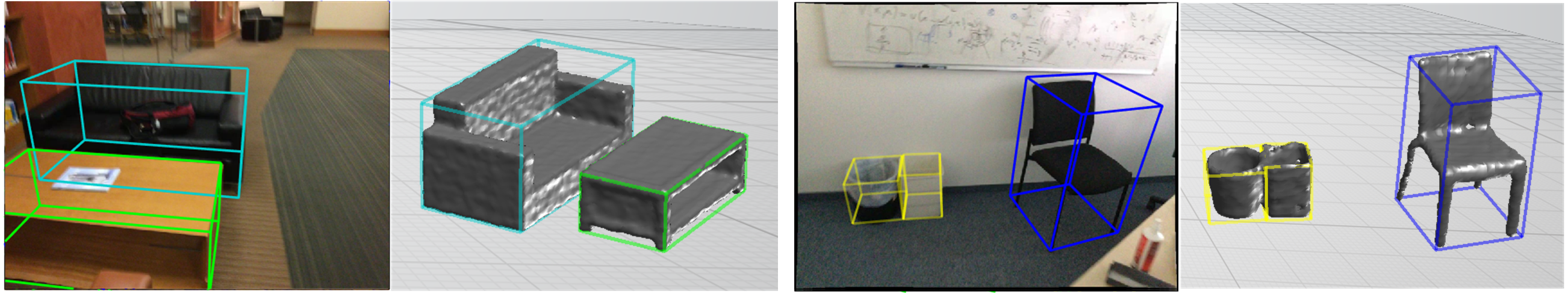}
\vspace{0mm}
\caption{ Given a single image as input, our proposed approach jointly predicts $3$D object bounding boxes and surfaces.}
  \label{fig:teaser}
  \vspace{0mm}
\end{figure}

In light of this, attempts like grid-based representation have been made for tasks such as rendering~\cite{sitzmann2019deepvoxels}, detection~\cite{roddick2018orthographic,reading2021categorical}, or reconstruction~\cite{guillard2020uclid}. OFT~\cite{roddick2018orthographic} proposes to sample and transform image features into a BEV grid representation, which enables holistic reasoning of the $3$D scene configuration. CaDDN~\cite{reading2021categorical} extends the BEV grid representation with a categorical depth prior, leading to higher $3$D detection accuracies. DeepVoxels~\cite{sitzmann2019deepvoxels} 
and UCLID-Net~\cite{guillard2020uclid} build voxel features by back-projecting $2$D features to $3$D space for respective rendering or single object reconstruction purposes. 
Inspired by this line of works,
we propose a novel voxel-based $3$D detection and reconstruction framework for predicting $3$D bounding boxes and surfaces of multiple objects from a single image (see Fig.~\ref{fig:teaser}).  

Specifically, we first divide a $3$D scene space into a regular grid of voxels. 
For each voxel, we assign $3$D features by sampling from the image plane via a $2$D-to-$3$D feature lifting operator and the known camera projection matrix.
As multiple voxels can be projected to the same position, this leads to similar features along the camera ray and increased difficulty for downstream tasks. 
To remedy this, we use a positional encoding strategy to make our voxel features position-aware and more discriminative. 
Based on the intermediate voxel features, we carefully devise our detection and reconstruction modules. 
For detection, we introduce a novel CenterNet-$3$D detector head. 
Instead of formulating the $3$D detection as $2$D keypoint detection problem as conventional CenterNet-based methods~\cite{zhou2019objects,engelmann2016joint}, each object is directly represented by its $3$D keypoint. 
Predicting a class-specific $3$D heatmap can show probabilities of $3$D object centers in the pre-defined voxel space, leading to improved $3$D center accuracy.
For reconstruction, we propose a multi-level shape representation with two components: coarse-level occupancy representation and fine-level local PCA-SDF representation. 
The coarse-level voxel grid represents the whole $3$D scene with continuous occupancy values. 
At a fine level, we represent the occupied voxels with a PCA-based signed distance function (SDF) by assuming that the local shapes of different voxels are similar either within an object instance, or across different objects.

In summary, the contributions of this work include:

 $\diamond$ We propose a novel voxel-based $3$D detection and reconstruction framework, which infers the $3$D locations and 3D surfaces for multiple object instances with only a $2$D image as input.

 $\diamond$ We present a novel CenterNet-$3$D detector, where each object is represented by its center point in a partitioned $3$D grid space. CenterNet-$3$D avoids estimating depth directly from image features, leading to increased detection performance.

 $\diamond$ We propose a novel local PCA-SDF shape representation, which provides finer reconstruction and order of magnitude faster inference than SOTA local implicit function methods like DeepLS~\cite{chabra2020deep}.

 $\diamond$ We demonstrate the superiority of our method in multiple object $3$D reconstruction and detection, as well as $3$D shape representation. We assemble a $3$D detection and reconstruction benchmark with $18,000$ real images, annotated with $3$D models and bounding boxes of $19$ object categories.


\Section{Related Work}

\Paragraph{$3$D Scene Understanding and Single Object Reconstruction.}
Tremendous efforts have been devoted to instance-level $3$D scene understanding~\cite{zeeshan2014cars,chabot2017deep,chen2016monocular,li2017deep,su2015render,tulsiani2015viewpoints,xiang2015data,zia2015towards,zhou2019objects,kundu20183d,brazil2019m3d} over the last decade. 
However, most of these approaches estimate object orientation~\cite{su2015render,tulsiani2015viewpoints,poirson2016fast} or $3$D bounding boxes~\cite{atoum2017monocular,chen2016monocular,chen20153d,xu2018multi,simonelli2019disentangling,liu2019deep,brazil2019m3d,choi2013understanding,huang2018cooperative,huang2018holistic}. 
Since describing objects with boxes only offers a coarse information of $3$D objects in images, the usage of $3$D models as shape priors can complement and enrich $3$D scene understanding.
Yet, scene understanding at the instance level remains challenging due to the large number of objects with various categories. 
With the substantial growth in the number of publicly available $3$D models, datasets such as ShapeNet~\cite{chang2015shapenet}
have allowed neural networks to train on the $3$D shape reconstruction task from single images~\cite{jiang2019disentangled,ranjan2018generating,liu20193d,bagautdinov2018modeling,dai2017shape,stutz2018learning}. To further leverage real-world images in $3$D modeling, as Liu \emph{et al.}~\cite{fully-understanding-generic-objects-modeling-segmentation-and-reconstruction} propose a semi-supervised learning framework for generic objects.
However, most of these methods estimate $3$D shapes in the object-centric coordinate system, which differs from the shape prediction of multiple instances at scene-level $3$D reconstruction -- the focus of our work.




\Paragraph{Multiple Object $3$D Reconstruction}
A common characteristic amongst aforementioned single object $3$D reconstruction approaches is that they usually treat objects as isolated geometries 
without considering the scene context, such as object locations, and instance-to-instance interactions. 
Recently, there is progress in multiple object $3$D reconstruction.  
$3$D-RCNN~\cite{kundu20183d} exploits the idea of using inverse graphics to map image regions to the $3$D shape and pose of object instances. 
The shape is represented by a simple linear subspace which limits its application for objects with large intra-class variability.  
Mesh R-CNN~\cite{meshrcnn} augments Mask R-CNN~\cite{he2017mask} with a mesh predictions branch that estimates a $3$D mesh for each detected in an image. 
Total$3$DUnderstanding~\cite{nie2020total3dunderstanding} presents a framework that predicts room layout, $3$D object bounding boxes, and meshes for all objects in an image based on the known $2$D bounding boxes.
However, these three methods first detect objects in the $2$D image, and then \emph{independently} produce their $3$D shapes with single object reconstruction modules. 
This could be problematic when $3$D boxes of objects intersect, such as a chair is pushed under a table.

Recently, CoReNet~\cite{popov2020corenet} performs multiple object reconstructions in a fixed $128^3$ voxel grid without recovering $3$D position information in the world space. 
Points2Objects~\cite{engelmann2020points} combines a $3$D object detector and shape retrieval to detect and reconstruct $3$D objects from an image. 
However, it suffers from two limitations: 1) Its CenterNet-based $3$D detector reasons $3$D boxes directly in the $2$D image domain, which is inherently challenging due to the lack of reliable depth cue.
2) Retrieval-based methods depend on the size and diversity of the pre-defined CAD model pool. 
Moreover, \cite{popov2020corenet,engelmann2020points} train on synthetic renderings, 
which limits their applicability to real-world scenarios.
Instead of relying on $2$D feature for $3$D detection or reconstruction, we propose to learn a geometry and context preserving voxel feature representation, which is well suited for $3$D detection and reconstruction. Moreover, we validate our method on real-world images from $19$ object categories.  




\Paragraph{Local Shape Priors}
Many neural architectures are proposed to  model $3$D objects via geometric representations, {\it e.g.}, point clouds~\cite{qi2017pointnet}, meshes~\cite{groueix2018atlasnet,wang2018pixel2mesh}, voxels~\cite{choy20163d,wu2017marrnet}, or implicit functions~\cite{chen2018learning,mescheder2018occupancy,park2019deepsdf,learning-implicit-functions-for-topology-varying-dense-3d-shape-correspondence}. 
Recently, neural implicit functions have demonstrated their effectiveness by encoding geometry in latent vectors and network weights, which parameterize surfaces through level-sets. 
Instead of an object-level representation, some follow-up works learn patch-level or primitive-level representations of surfaces, {\it e.g.}, PatchNet~\cite{genova2019learning}, CvxNet~\cite{deng2020cvxnet}, BSP-Net~\cite{chen2020bsp}. 
To further leverage local geometric priors, another line of works learn implicit geometry on sparse regular~\cite{jiang2020local,takikawa2021neural} or $3$D voxel grids~\cite{liu2020neural,chabra2020deep}. 
A latent code of each voxel is responsible for representing implicit geometry in a small neighborhood, enabling fine-grained reconstruction. 
However, these methods often suffer from inefficient inference as each point needs a forward pass through of the implicit function network. 
Instead, we build a local PCA-SDF shape representation, which represents each local shape as a linear combination of implicit volumetric prototypes, leading to finer details and order of magnitude faster inference than prior works. 
Similar eigenanalysis of SDF has been applied for either global shape representation~\cite{michalkiewicz2020simple} or geometry compression~\cite{ricao2017compressed,tang2018real}. 
However, none of them develop their algorithms from our perspective of local shape priors, which is motivated from the assumption that local shapes at the voxel level share similarities. 



\Section{Methodology}

\begin{figure}[t!]
\centering
\includegraphics[trim=0 0 0 0,clip,width=13.8cm]{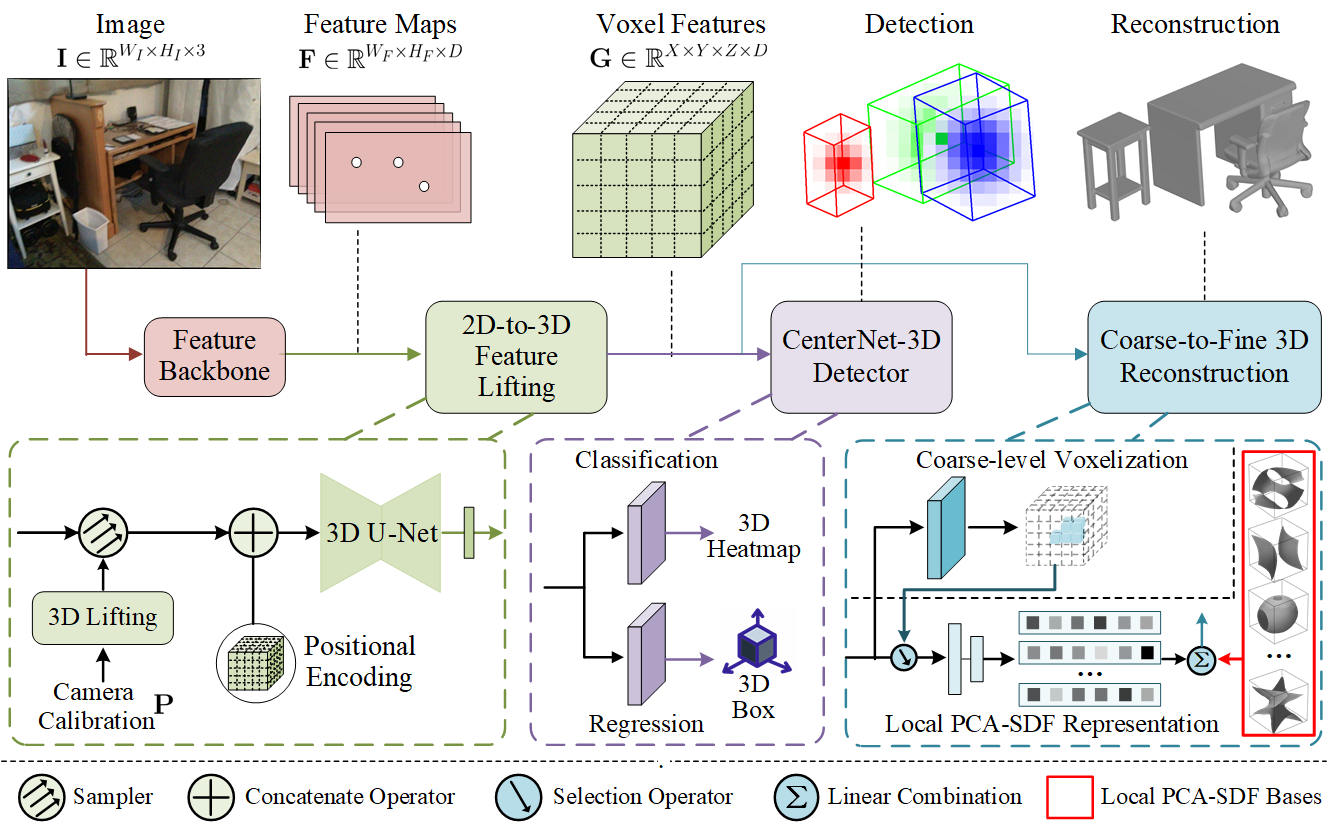}
\vspace{0mm}
\caption{\textbf{Overview of our approach.} The proposed joint framework is composed of three key modules: $3$D voxel feature learning (consists of feature backbone and $2$D-to-$3$D feature lifting), CenterNet-$3$D detector, and coarse-to-fine $3$D reconstruction. $2$D feature maps are first generated from input image $\mathbf{I}$, which are back-projected into voxel features $\mathbf{G}$ using a known camera projection matrix $\mathbf{P}$. The voxel features serve for our novel $3$D object detection and reconstruction.}
  \label{fig:overview}
  \vspace{-2mm}
\end{figure}

We illustrate the overall architecture of our method in Fig.~\ref{fig:overview}, which consists of three key modules:  i) $3$D voxel feature learning; ii) CenterNet-$3$D detector; and iii) Coarse-to-fine $3$D reconstruction. 
\SubSection{3D Voxel Feature Learning}
Our network learns to produce a compact $3$D voxel feature representation of the image with \emph{complementary} supervision from both $3$D detection and reconstruction, which enables rich $3$D context and geometric information and allows the two tasks to benefit each other.
%
We first define a $3$D grid $\mathbf{V}\in\mathbb{R}^{X \times Y \times Z}$ by partitioning of the scene space into voxels $\mathbf{V}_{i}$ with a voxel size of $r$. 
The $3$D grid size, $Xr \times Yr \times Zr$, is set based on the minimum volume of the scenes in the width, height and length dimensions, that can encompass all annotated instances in the dataset. 


\Paragraph{Feature Extraction.} We utilize a convolutional feature extractor to generate a hierarchy of multi-scale $2$D feature maps. Specifically, the input to the feature extraction network is a RGB image $\mathbf{I}\in \mathbb{R}^{W_{I}\times H_{I}\times 3}$, where $W_{I}$ and $H_{I}$ are the image size. 
The convolutions are followed by down scaling the input, creating growing receptive fields and resulting in $D$-channel multi-scale $2$D feature maps $\mathbf{F}\in \mathbb{R}^{W_{F}\times H_{F}\times D}$. 
$W_{F}$ and $H_{F}$ are the width and height of feature $\mathbf{F}$. 

\Paragraph{Lifting 2D Features to 3D.} The lifting layer back-projects $2$D feature maps $\mathbf{F}$ into $3$D voxel space, resulting in \emph{initial} $3$D voxel features $\mathbf{G}$. Formally, the objective of the lifting operator is to populate the $3$D voxel features $\mathbf{G}(x,y,z)$ with the projected $2$D features $\mathbf{F}\{(u,v)\}$, where $\{\cdot\}$ denotes bilinear interpolation on the $2$D feature maps. 
We assume a full perspective camera model. 
Any voxel center $(x,y,z)$ can be projected to image plane via a camera projection matrix $\mathbf{P}\in \mathbb{R}^{3\times 4}$: $[u\cdot d, v\cdot d, d]^{T}=\mathbf{P}[x,y,z,1]^{T}$. Here $u$ and $v$ are the $2$D position of the projection and $d$ is its depth from the camera. 
The resulting voxel features $\mathbf{G}\in \mathbb{R}^{X\times Y\times Z\times D}$ provide a scene representation that is free from the effects of perspective projection.

\Paragraph{3D Voxel Features Aggregation.}
The lifting mechanism we use is similar to the one in~\cite{roddick2018orthographic,guillard2020uclid,sitzmann2019deepvoxels}, which has a major weakness that all voxels along a camera ray will receive the same $2$D feature. 
This feature smearing issue increases the difficulties of $3$D detection and reconstruction. 
To mitigate this issue, we propose to employ the \emph{positional encoding} (PE)~\cite{vaswani2017attention} strategy that adds $3$D voxel center position to the voxel features: $\mathbb{R}^{X\times Y\times Z\times D} \xrightarrow[]{PE} \mathbb{R}^{X\times Y\times Z\times (D+3)} $, which helps the voxel features to be more discriminative and position-embedded. 
We further utilize a $3$D convolutional hourglass (U-Net) network~\cite{cciccek20163d}, comprised of a series of down- and upsampling convolutions with skip connections, to integrate both local and global information. The final voxel features are thus $\mathbf{G}: \mathbb{R}^{X\times Y\times Z\times (D+3)} \xrightarrow[]{U-Net} \mathbb{R}^{X\times Y\times Z\times (D+3)} $.
This voxel features serve as the cornerstone for the downstream tasks of $3$D detection and reconstruction.
\SubSection{Monocular CenterNet-$3$D Detector}
Conventional CenterNet-based~\cite{zhou2019objects} monocular $3$D detection methods such as Points2Objects~\cite{engelmann2020points} formulate $3$D detection as a projected \emph{$2$D keypoint} detection problem.
In contrast, we propose a novel CenterNet-$3$D detection head, where each object is directly represented by its \emph{$3$D keypoint}. 
%
Then $3$D properties such as object size and orientation can be intuitively inferred from the $3$D voxel features at the center location by a regression branch. 
Compared to~\cite{engelmann2020points}, 
CenterNet-$3$D avoids estimating depth values of $3$D boxes directly from $2$D features, leading to improved detection accuracy.  

\Paragraph{3D Keypoint Branch.}
The $3$D keypoint branch takes the voxel features $\mathbf{G}$ as input and predicts a \emph{$3$D heatmap} $\mathcal{Y}\in \mathbb{R}^{X\times Y \times Z \times C}$ (see Fig.~\ref{fig:overview}), where $C$ is the number of object categories. 
The values of each voxel in $\mathcal{Y}$ indicates how likely the $3$D centroid of a certain object category exists at the voxel center. 
By computing the local maxima and filtering via a threshold, we obtain a preliminary estimation of the $3$D centroids, denoted as $\tilde{\mathbf{c}}_{3d}=[x_c,y_c,z_c]^T$. 
To remedy the discretization error of voxels, the regression branch additionally predicts a local offset to $\tilde{\mathbf{c}}_{3d}$, which is discussed next. 

\Paragraph{Regression Branch.} The regression branch predicts the essential properties to construct a $3$D bounding box for each voxel of the $3$D heatmap. We parameterize a $3$D box as the prior works~\cite{huang2018cooperative,nie2020total3dunderstanding} and set up the world system located at the camera center with its vertical ($y$-) axis perpendicular to the floor and its forward ($z$-) axis toward the camera, such as the pitch and roll angles could be included in the camera pose.
Specifically the $3$D box is encoded as a $8$-tuple $\tau=[\delta_{x_{c}}, \delta_{y_{c}}, \delta_{z_{c}}, \delta_h, \delta_w, \delta_l, \sin \theta, \cos \theta]$. Here $\Delta \mathbf{c}_{3d}=[\delta_{x_{c}}, \delta_{y_{c}}, \delta_{z_{c}}]^{T}$ denotes the $3$D center offset compensating voxel discretization.
$[l,h,w]^{T}=[\bar{l}\cdot e^{\delta_{l}}, \bar{h}\cdot e^{\delta_{h}}, \bar{w}\cdot e^{\delta_{w}}]^{T}$ represents the object size, where $[\bar{l}, \bar{h}, \bar{w}]^{T}$ is a pre-calculated category-wise average box size, $[ \delta_h, \delta_w, \delta_l]$ represents the corresponding transformations. 
$\theta$ denotes the rotation angle around $y$-axis. 
Here the network estimates the vectorial representation of rotation angle $\theta$~\cite{liu2020smoke}. The output size of the regression branch is thus $X\times Y\times Z\times 8$.
Given the outputs of keypoint and regression branches, the $3$D bounding box $\mathcal{B}\in\mathbb{R}^{3\times8}$ can be restored as $8$ corners: 
\begin{equation}
    \mathcal{B} = R_\theta\begin{bmatrix} \pm l/2\\ \pm h/2\\ \pm w/2\end{bmatrix} + \mathbf{c}_{3d}, \quad \mathbf{c}_{3d} = \tilde{\mathbf{c}}_{3d}+\Delta \mathbf{c}_{3d},
\end{equation}
where $R_\theta\in \mathbb{R}^{3\times3}$ is the rotation matrix.


\begin{figure}[t!]
\centering    
\subfigure[]{\label{fig:shape_representation_a}\includegraphics[trim=0 0 0 0,clip, height=32mm]{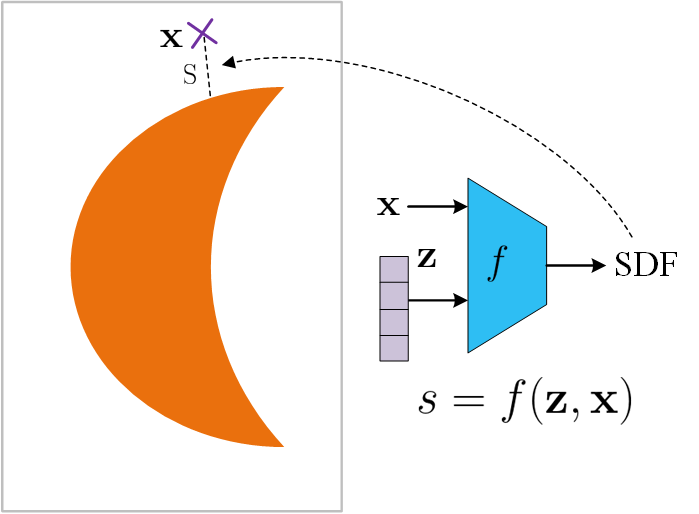}} 
\hspace{1mm}
\subfigure[]{\label{fig:shape_representation_b}\includegraphics[trim=0 0 0 0,clip,height=32mm]{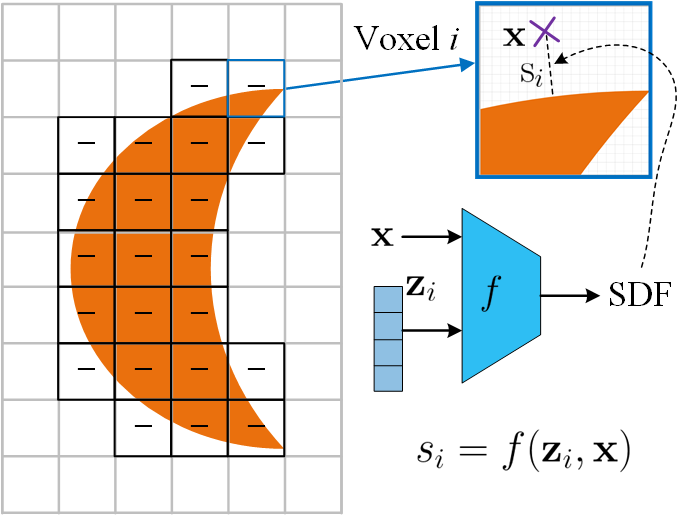}}
\hspace{1mm}
\subfigure[]{\label{fig:shape_representation_c}\includegraphics[trim=0 0 0 0,clip,height=32mm]{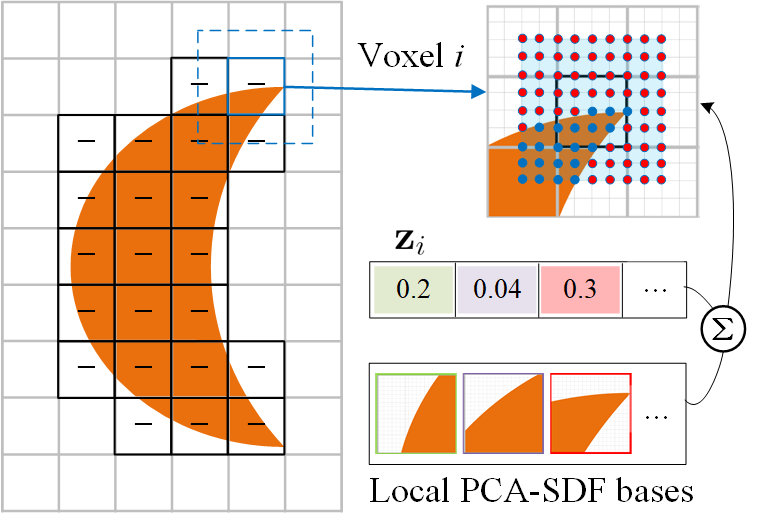}}
\vspace{0mm}
\caption{ $2$D examples of (a) DeepSDF~\cite{park2019deepsdf}, (b) DeepLS~\cite{chabra2020deep}, and (c) our local PCA-SDF shape representation. DeepSDF describes the surfaces with global shape codes. The SDF function $f$ in DeepLS outputs a scalar value conditional on the local latent code $\mathbf{z}_{i}$ and local coordinate $\mathbf{x}$. However, its inference is computationally expensive since it requires forward pass through $f$ for every $\mathbf{x}$. 
Our shape representation consists of coarse-level voxelization and fine-level local PCA-SDF. The coarse-level voxelization holistically represents the whole 3D surface with binary values. 
To further represent fine-level surfaces, we propose a novel local PCA-SDF model, representing any occupied voxel as a linear combination of regular SDF function bases, which enables a more efficient and accurate representation than DeepLS.}

\vspace{-2mm}
\label{fig:shape_representation}
\end{figure}

\SubSection{Coarse-to-Fine $3$D Reconstruction}
Our reconstruction module is based on a coarse-to-fine shape representation, which consists of two components: coarse-level voxelization and fine-level local PCA-SDF. 



\Paragraph{Coarse-Level Voxelization.}
Based on the extracted $3$D voxel features $\mathbf{G}$, we first estimate a coarse-level voxelization $\tilde{\mathcal{V}}$ by a specific branch. 
The coarse-level voxelization holistically represents the whole $3$D surface with binary occupancy values, where the unoccupied voxels cover ``air" in the scene, and occupied voxels can be either fully occupied ones inside the object, or voxels intersecting with the object's surface.
For the occupied voxels of both types, we further reconstruct a fine-level local shape via the local PCA-SDF.

\Paragraph{Local PCA-SDF Shape Representation.}
Recent works such as~DeepSDF~\cite{park2019deepsdf} aims to learn global implicit functions to represent shapes (see Fig.~\ref{fig:shape_representation_a}). However, representing the entire objects with a single latent code often results in loss of details, which limits its application for scene-level object reconstruction.
%
~DeepLS~\cite{chabra2020deep} represents $3$D surfaces by a set of independent latent codes on a regular grid (see Fig.~\ref{fig:shape_representation_b}). Each latent code $\mathbf{z}_{i}$, concatenated with any point location $\mathbf{x}$, can be decoded into a SDF value $\textup{s}_{i}$ by the learned implicit network $f$: $s_{i}=f(\mathbf{z}_{i}, \mathbf{x})$. 
However, this shape representation has two limitations.
i) Inference is inefficient (in the order of seconds) since every point of a test voxel (\emph{e.g.,} $256^3$ points) is required to be sent to $f$ for SDF calculation, making it unsuitable for real-time applications. 
ii) Our key observation is that local voxels, either within an object instance or across different categories, share similar local shapes, {\it e.g.}, voxels across a table's surface all have planar shapes. However, DeepLS treats voxels as independent training samples, without fully leveraging such local shape priors in training.  
To address these issues, we propose a novel local PCA-SDF shape representation, which represents each voxel shape as a linear combination of a set of implicit volumetric prototypes, leading to significantly finer reconstruction and $10\times$ faster inference speed than~DeepLS (see Tab.~\ref{tab:shape_representation}).  

Formally, as shown in Fig.~\ref{fig:shape_representation_c}, for each occupied voxel $\mathbf{V}$, we define a regular lattice $\mathbf{q}\in \mathbb{R}^{k\times k\times k \times 3}$ 
and compute their SDFs $\mathbf{s}\in \mathbb{R}^{k\times k\times k \times 1}$ toward the surface. 
By collecting SDFs of $N_{S}$ occupied voxels from the training surfaces, we apply Principal Component Analysis (PCA) to find $l_{B}$ ($l_{B}<<k^3$) local shape bases, $\mathcal{S}_{B}\in \mathbb{R}^{ k\times k\times k \times l_{b}}$. As such, given the learned $\mathcal{S}_{B}$ and the latent code $\mathbf{z}_{i}$, any local shape $\mathcal{S}_{i}$ of the underlying surface can be implicitly represented by $\mathcal{S}_{i}=\mathcal{S}_{B} \mathbf{z}_{i}$. The latent code $ \mathbf{z}_{i}$ for $\mathcal{S}_{i}$ can be generated by the corresponding voxel feature $\mathbf{G}_{i}$ via $\mathbf{z}_{i}= \textup{MLP}(\mathbf{G}_{i})$. MLP is a mapping network, implemented with two fully-connected layers.
By combining the contributions of all the occupied voxels, we can infer a global iso-surface from the SDF field. 
%
%
Similar to DeepLS~\cite{chabra2020deep}, we apply a $1.5$ times receptive field strategy to mitigate the inconsistent surface predictions at the voxel boundaries (Fig.~\ref{fig:shape_representation_c}). Accordingly, during inference, we could apply average pooling to combine the SDF values for the boundary area. It is worth mentioning that our local PCA-SDF also allows reconstruction at resolutions higher than the one used during training by simply applying trilinear interpolation on the learned local shape bases $\mathcal{S}_{B}$.

\SubSection{Loss Functions and Implementation Details}
The training data of one example consists of RGB image $\mathbf{I}$, ground-truth $3$D bounding boxes $\mathcal{B}^{*}$ of objects, coarse-level voxelization $\mathcal{V}^{*}$, and a set of regular SDF pairs $\{(\textup{idx}_{j}, \mathbf{s}_{j})\}_{j=1}^{K}$ sampled from the surface. Here, each set of SDFs  $\mathbf{s}_{j}\in \mathbb{R}^{k\times k\times k}$, $\textup{idx}_{j}$ is the $\textup{idx}_{j}$-th voxel of the holistic grid. During training, we jointly optimize the parameters of $2$D feature extraction network, $3$D U-Net, detection and reconstruction modules by minimizing three losses: $3$D keypoint classification loss $\mathcal{L}_{cls}$, regression loss $\mathcal{L}_{reg}$, and $3$D reconstruction loss $\mathcal{L}_{recon}$, \emph{i.e.,}

\begin{equation}
    \mathcal{L} = \mathcal{L}_{cls} + \mathcal{L}_{reg} + \mathcal{L}_{recon}.
\end{equation}

\Paragraph{Loss Functions.} We generate the target heatmaps $\mathcal{Y}^{*}$ by splatting the ground truth $3$D center points using a Gaussian kernel (please refer to \textbf{Supp} for details). If two Gaussians of the same class overlap, we take the element-wise maximum. The $3$D keypoint branch is trained with a penalty-reduced focal loss~\cite{lin2017focal,zhou2019objects} in a point-wise manner on the $3$D heatmap.

\begin{equation}
    \mathcal{L}_{cls} = \frac{-1}{N} \sum_{xyzc} \left\{\begin{matrix} (1-\mathcal{Y}_{xyzc})^\mu \textup{log}(\mathcal{Y}_{xyzc}) & \textup{if} \quad \mathcal{Y}^*_{xyzc}=1
    \vspace{3mm}
\\ 
(1-\mathcal{Y}^*_{xyzc})^{\sigma} (\mathcal{Y}_{xyzc})^\mu \textup{log}(1-\mathcal{Y}_{xyzc}) &  \textup{otherwise}
\end{matrix}\right.  
\end{equation}
where $N$ is the number of objects per image, $\mu=2$ and $\sigma=4$ are hyper-parameters of the focal loss.

We define the $3$D bounding box regression loss as the $L_{1}$ distance between the predicted transform $\mathcal{B}$ and the ground truth $\mathcal{B}^*$:
    $\mathcal{L}_{reg} =  \frac{1}{N} || \mathcal{B} - \mathcal{B}^*||_{1}$.
The reconstruction loss consists of cross-entropy classification loss $\mathcal{L}_{v}$ for coarse-level voxelization and fine-level SDF regression loss.
\begin{equation}
    \mathcal{L}_{recon} = \mathcal{L}_{v}(\mathcal{V}, \mathcal{V}^*) + \sum_{j}^{K}|| \mathcal{S}_{B}\mathbf{z}_{j}-\mathbf{s}_{j}||^{2}_{2}, \quad  \mathbf{z}_{j}=\textup{MLP} (\mathbf{G}_{\textup{idx}_{j}}).
\end{equation}

\Paragraph{Implementation Detail.}
We use a ResNet-$34$ network as our $2$D feature extractor. 
We extract features immediately before the final three downsampling layers, resulting in a set of feature maps at $1/8$, $1/16$, $1/32$ scales of the input resolution. 
We then resize them to the original size of the input images via bilinear interpolation. 
Convolutional layers with $1\times 1$ kernels are used to map these feature maps to a common channel size of $64$. 
The set of feature maps is summarized as $\mathbf{F}$ before passing to $3$D lifting layers. 
For the main experiments, we train our models with a batch size of $8$ on a GTX $1080$Ti GPU for $200$ epochs. The learning rate is set at $5\times10^{-4}$ and drops at $50$ and $100$ epochs by a factor of $10$. For more details, please refer to Sec.~\ref{sec:exp} or \textbf{Supp}. 


\Section{Experiments}
\label{sec:exp}

\begin{table}
  \caption{ \textbf{Multiple object reconstruction comparison.} We report per-class and mean IoU over all classes, and class-agnostic global IoU on $128^3$ voxel grid.}
  \centering
  \vspace{0mm}
  \resizebox{0.9\linewidth}{!}{
  \begin{tabular}{l |c c c c c c | c c || c c}
    \toprule
    \multirow{2}{*}{Method} & 
    \multicolumn{8}{c||}{ShapeNet-triplets}   &    \multicolumn{2}{c}{ShapeNet-pairs}           \\
    \cmidrule(r){2-9}
    \cmidrule(r){10-11}
    &bottle & bowl & chair & mug & sofa & table & mean & global & mean & global \\
    \midrule
    CoReNet~\cite{popov2020corenet} & $61.8$  & $36.2$ & $30.1$ & $48.0$ & $52.9$ & $34.8$ & $43.9$ &  $49.8$  & $43.1$ &  $52.7$   \\ 
    Points2Objects~\cite{engelmann2020points} &  $\textbf{63.5}$  & $30.2$ & $18.9$ & $41.5$ & $44.5$ & $19.8$ & $36.4$ &  $44.7$  & $-$ &  $-$   \\ \hline
    Proposed     & $63.3$ & $\textbf{38.5}$ & $\textbf{31.8}$ & $\textbf{51.7}$  & $\textbf{54.3}$ & $\textbf{36.1}$  & $\textbf{46.0}$ & $\textbf{52.3}$ & $\textbf{46.7}$ & $\textbf{55.1}$   \\ 
    \bottomrule
  \end{tabular}
  }
  \label{tab:reconstruction}
  \vspace{0mm}
\end{table}

\begin{figure}[t]
\vspace{0mm}
\resizebox{1\linewidth}{!}{
\begin{tabular}{@{\hspace{-0.01cm}} c @{\hspace{-0.01cm}} c @{\hspace{-0.01cm}}}
   \rotatebox[origin=c]{90}{\small Input} &  \raisebox{-.5\height}{\includegraphics[scale=0.25]{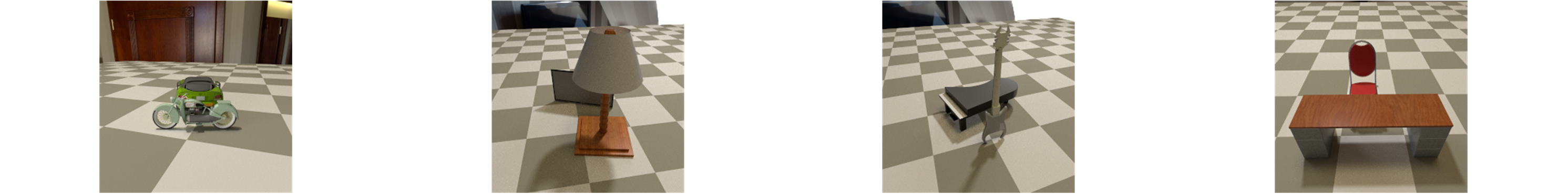}} \\
\rotatebox[origin=c]{90}{\small CoReNet}  &
\raisebox{-.5\height}{\includegraphics[scale=0.25]{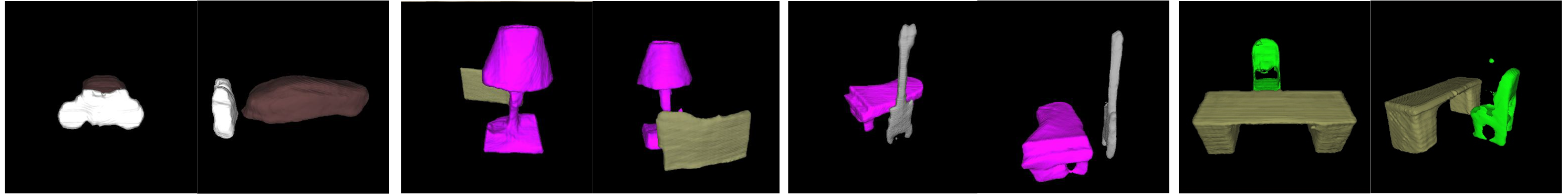}} \\
\rotatebox[origin=c]{90}{\small Proposed}  & \raisebox{-.5\height}{\includegraphics[scale=0.25]{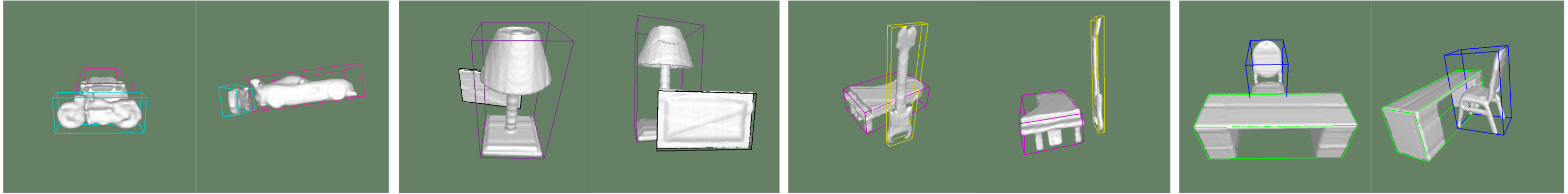}} \\
\end{tabular}
}
 \vspace{0mm}
\caption{ \textbf{Qualitative results on ShapeNet-triplets dataset.} We compare to CoReNet~\cite{popov2020corenet} in two different viewpoints. Our model more accurately reconstructs details and hallucinates occluded parts. } 
\label{fig:shapenet_reconstruction}
  \vspace{0mm}
\end{figure}

\SubSection{Multiple Object Detection and Reconstruction on ShapeNet-pairs and -triplets}
\label{sec:shapenet-recon}
\Paragraph{Datasets.}
Following the experimental setting of CoReNet~\cite{popov2020corenet} and Points2Objects~\cite{engelmann2020points}, we evaluate multiple object detection and reconstruction on \emph{ShapeNet-pairs} and \emph{ShapeNet-triplets} datasets~\cite{popov2020corenet}. 
These datasets contain $256\times 256$ photorealistic renderings of either pairs or triplets of ShapeNet~\cite{chang2015shapenet} objects placed on a ground plane with random scale, rotation, and camera viewpoint. 
The \emph{ShapeNet-pairs} has several pairs of object classes: bed-pillow, bottle-bowl, bottle-mug, chair-table, display-lamp, guitar-piano, motorcycle-car, which contains $365,600$ images on trainval and $91,200$ on test. The \emph{ShapeNet-triplets} is with bottle-bowl-mug and chair-sofa-table, which includes $91,400$ on trainval and $22,000$ on test.

\Paragraph{Experimental Settings.}
In this experiment, we set a voxel grid of size $X{\times} Y{\times} Z$ = $40{\times} 25 {\times} 25$ $(r=0.1)$, which is sufficient to enclose all objects in the datasets. We randomly select $200$ surfaces from the training set to generate $N_{S}\approx 200,000$ occupied voxels. In this experiment, we set $k=11$, $l_{B}=64$, $C=6$ (ShapeNet-triplets) or $C=13$ (ShapeNet-pairs). 
We compare with SOTA methods for multiple object reconstruction: CoReNet~\cite{popov2020corenet} and Points2Objects~\cite{engelmann2020points}. 
As CoReNet doesn't perform $3$D detection, we only compare with Points2Objects on detection. 
%
Following~\cite{engelmann2020points}, we use mean average precision (mAP) as the detection metric with $3$D box intersection-over-union (IoU) thresholds $0.25$ and $0.5$. 
Following~\cite{popov2020corenet,engelmann2020points}, the metric for reconstruction is  IoU on a $128^3$ voxel grid.

\Paragraph{Results.}
We first report the $3$D detection results. Our method achieves a higher detection accuracy than Points2Objects~\cite{engelmann2020points}: $\textbf{51.5}\%$ {\it vs.}~$48.6\%$ (threshold$@0.5$) and $\textbf{80.3}\%$ {\it vs.}~$77.2\%$ (threshold$@0.25$), which demonstrates that voxel features perform better than image-based features for monocular $3$D detection.
For reconstruction, we report the mean over the per-class IoU, as well as the global IoU of all object instances within a scene, which does not concern predicted class labels. As compared in Tab.~\ref{tab:reconstruction}, our method significantly outperforms two SOTA baselines on both datasets. 
On ShapeNet-triplets, our method achieves relative $5.0\%$ global IoUs gains while $4.8\%$ mean IoUs gains, which indicates that our model performs well on reconstructing the overall shapes of objects.
%
Qualitative results of detection and reconstruction are shown in Fig.~\ref{fig:shapenet_reconstruction}.

\begin{figure}[t]
\centering    
\subfigure[]{\label{fig:pix3d} \includegraphics[trim=0 4 0 0,clip,height=39mm]{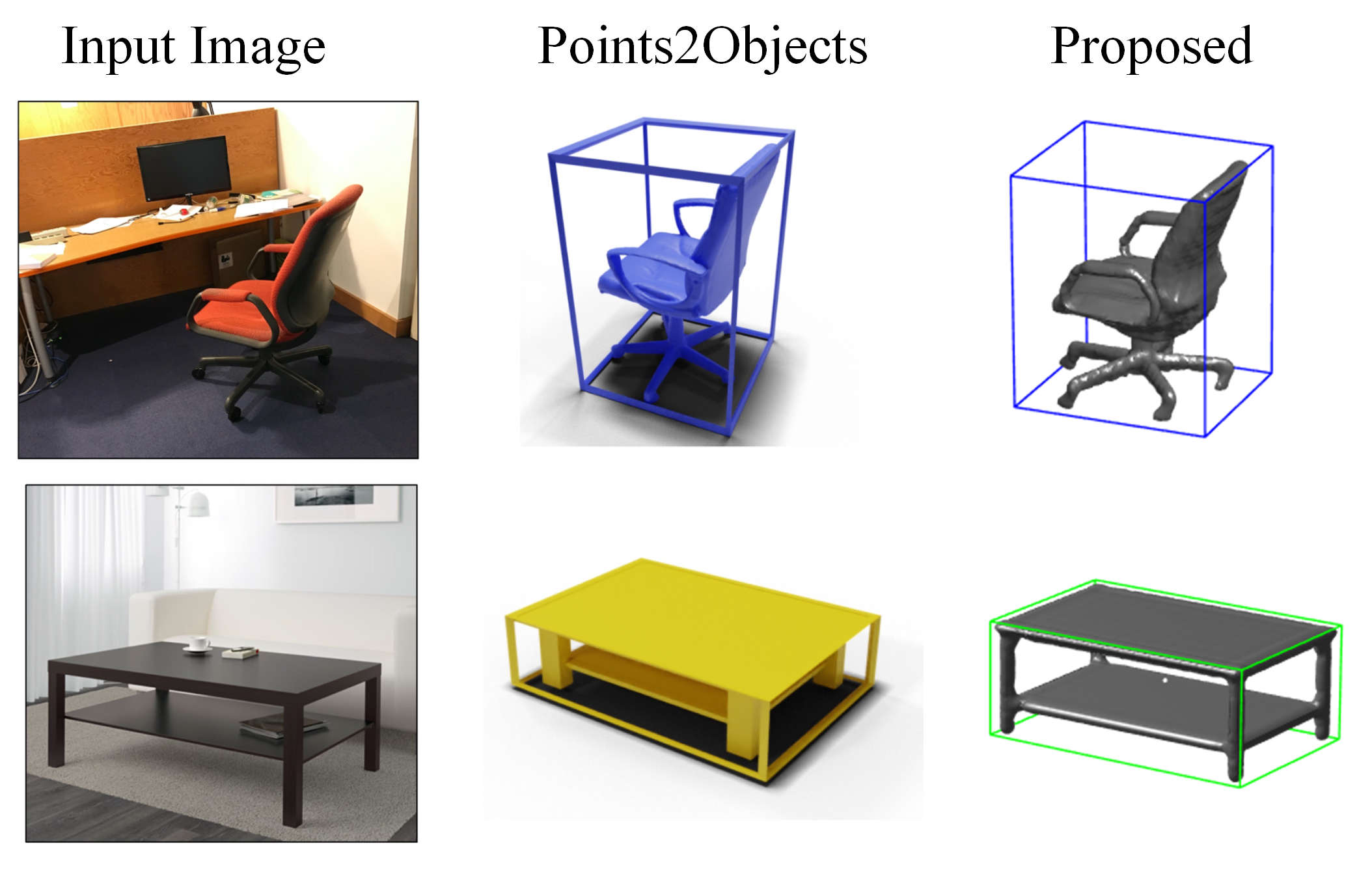}} 
\hspace{2mm}
\subfigure[]{\label{fig:pca}\includegraphics[trim=0 0 0 0,clip,height=39mm]{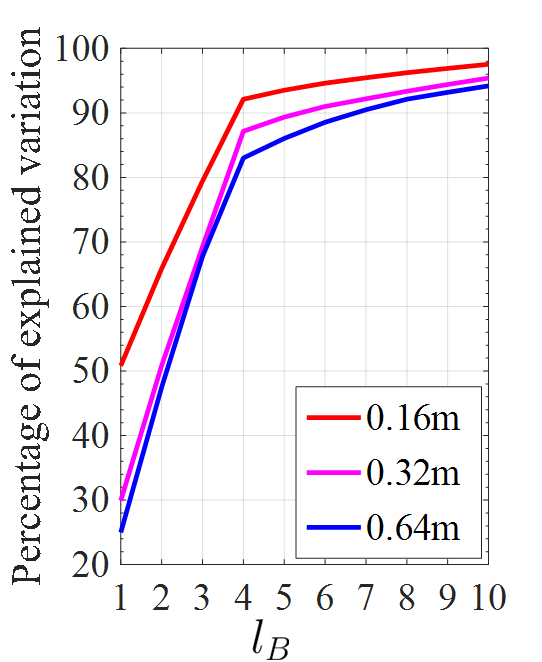}}
\hspace{-1mm}
\subfigure[]{\label{fig:sensitive}\includegraphics[trim=0 0 0 0,clip,height=39mm]{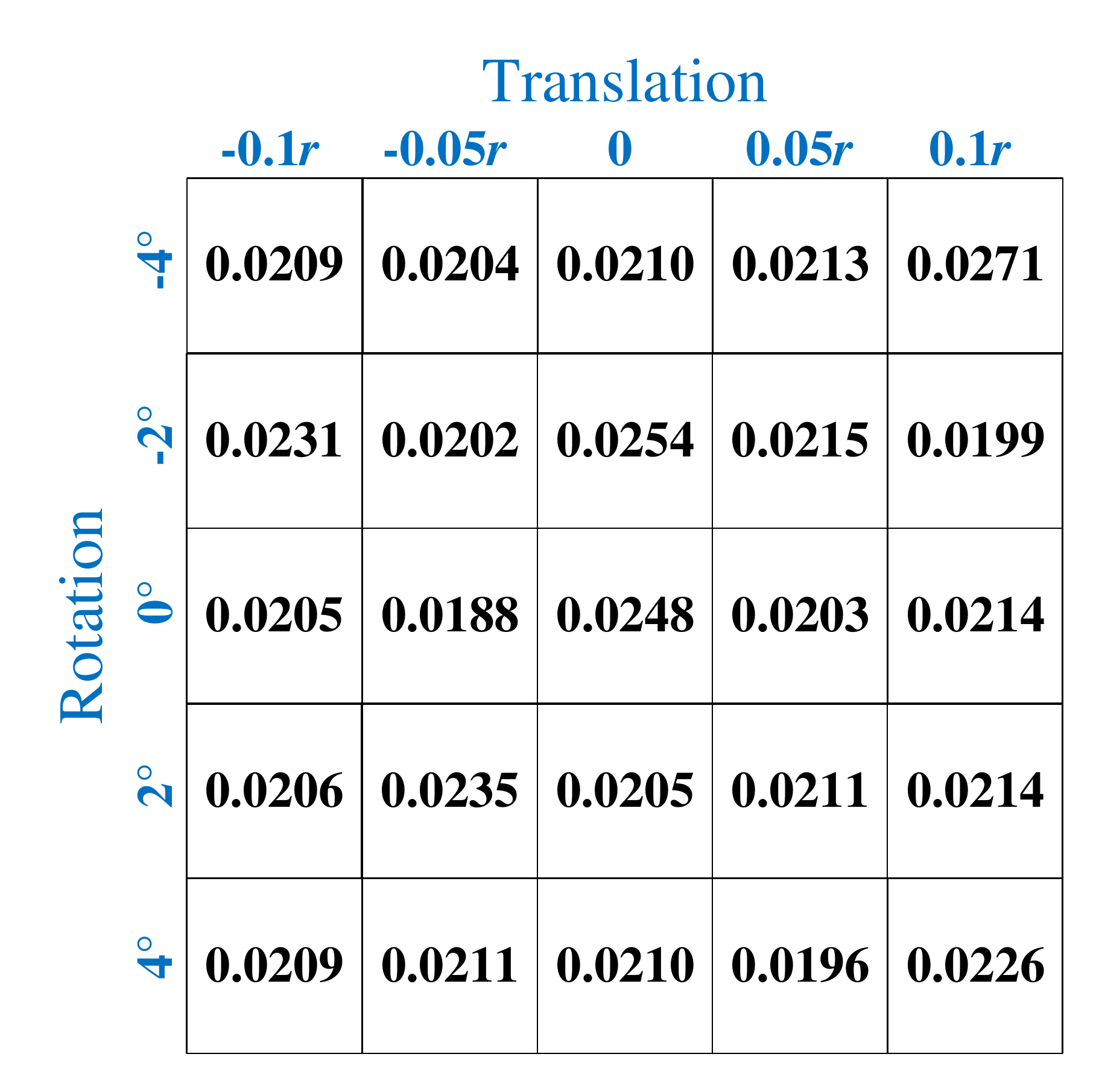}}
\vspace{0mm}
\caption{ (a) Qualitative comparison on Pix$3$D. Our reconstructions closely match objects' genuine shape, \emph{e.g.,} the table legs and chair arms.
(b) Explained variation of our PCA-SDF representation with three voxel sizes $r$.
(c) Reconstruction errors (Chamfer Distance-$L2$) of PCA-SDF w.r.t.~translated and rotated $3$D shapes.}
\vspace{0mm}
\label{fig:mixure_figure}
\end{figure}

\SubSection{Single Object Reconstruction on Pix3D}
 We further compare to CoReNet~\cite{popov2020corenet} and Points2Objects~\cite{engelmann2020points} on the real image database, Pix$3$D~\cite{sun2018pix3d} in the same protocol (splits $S_1$ and $S_2$) as in~\cite{meshrcnn}.
 In this experiment, we train our model with the same experimental setting and the same pre-computed PCA-SDF bases as in Sec.~\ref{sec:shapenet-recon}.
 On average IoU over all $9$ object classes, we achieve $\textbf{38.6}\%$ $vs.$ $34.1\%$ (CoReNet) $vs.$ $33.3\%$ (Points2Objects) on $S_1$ and $\textbf{28.6}\%$ $vs.$ $26.3\%$ (CoReNet) $vs.$ $23.6\%$ (Points2Objects) on $S_2$. The results demonstrate that our approach improves over baselines on real images.
Qualitative results are shown in Fig.~\ref{fig:pix3d}.

\begin{figure}[t]
\begin{center}
\resizebox{1\linewidth}{!}{
\begin{tabular}{@{\hspace{-0.005cm}} c @{\hspace{-0.005cm}} c @{\hspace{-0.005cm}}}
 \rotatebox[origin=c]{90}{\small Input} & \raisebox{-.5\height}{\includegraphics[scale=0.25]{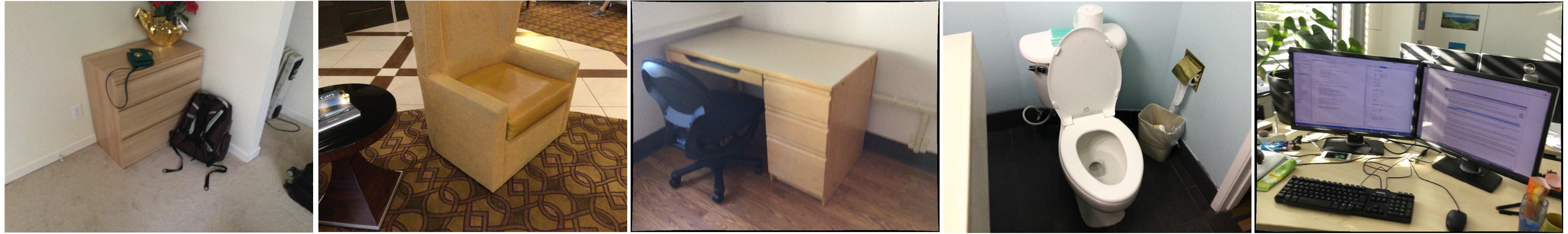}} \\
\rotatebox[origin=c]{90}{\small CoReNet}  & \raisebox{-.5\height}{\includegraphics[scale=0.25]{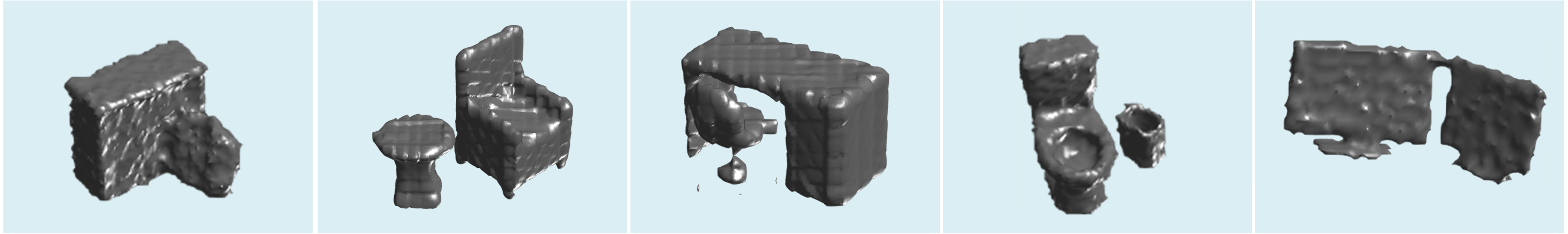}} \\
\rotatebox[origin=c]{90}{\small Proposed} & \raisebox{-.5\height}{\includegraphics[scale=0.25]{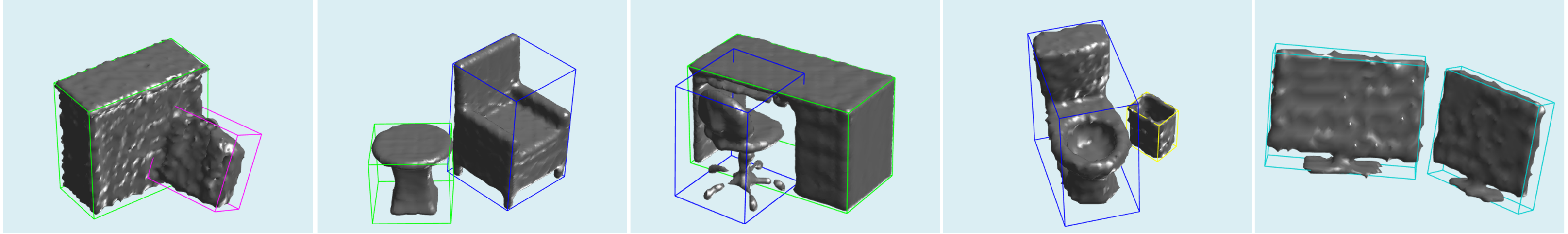}} \\
\rotatebox[origin=c]{90}{\small Ground-truth} & \raisebox{-.5\height}{\includegraphics[scale=0.25]{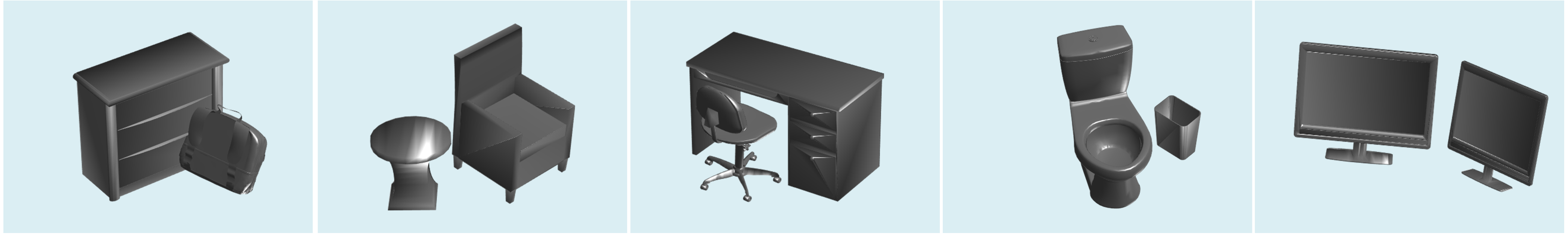}} \\
\end{tabular}
}
\vspace{0mm}
\caption{ \textbf{Qualitative results on real images from ScanNet-MDR.} Our reconstructions closely match the objects than CoReNet~\cite{popov2020corenet}. Moreover, our method performs better for reconstruction of the truncated objects.} 
\label{fig:scannet_reconstruction}
\end{center}
\vspace{-3mm}
\end{figure}

\begin{table}[t]
\begin{minipage}[t]{0.55\linewidth}
\centering
\caption{ Comparisons of $3$D object detection and reconstruction on ScanNet-MDR dataset. [Key: \textcolor{blue}{\textcircled{\normalsize {1}}}=CenterNet, \textcolor{blue}{\textcircled{\normalsize{2}}}=CenterNet-$3$D], \textcolor{red}{\textcircled{\normalsize{1}}}=DeepLS, \textcolor{red}{\textcircled{\normalsize{2}}}=Local PCA-SDF]}
\vspace{0mm}
\centering
\resizebox{1\linewidth}{!}{
  \begin{tabular}{l |c c| c c || c |c }
    \toprule
  \multirow{2}{*}{Method} & \multicolumn{2}{c|}{Detection}  & \multicolumn{2}{c||}{Recon.} &  \multicolumn{2}{c}{Evaluation}                  \\
    \cmidrule(r){2-3} \cmidrule(r){4-5} \cmidrule(r){6-7} & \textcolor{blue}{\textcircled{\normalsize {1}}} & \textcolor{blue}{\textcircled{\normalsize{2}}}
    & \textcolor{red}{\textcircled{\normalsize{1}}} & \textcolor{red}{\textcircled{\normalsize{2}}} & mAP (@$0.15$) & IoU  \\
    \midrule
 CoReNet~\cite{popov2020corenet} &  & &$-$ &  $-$  & $-$ & $35.2$ \\ \hline   

   Proposed-$1$  &  &  &  & \greencheck & $-$ & $36.5$ \\
   Proposed-$2$  &  & \greencheck &  & & $20.7$ & $-$ \\ 
   Proposed-$3$  &  \greencheck  &  &  & \greencheck & $19.5$ & $36.9$ \\
   Proposed-$4$  &  &  \greencheck  &  \greencheck & & $20.0$ & $35.4$\\ \hline
     Proposed-w/o PE      &  & \greencheck &  & \greencheck &  $\textbf{21.2}$ & $\textbf{37.2}$ \\ 
    Proposed      &  & \greencheck &  & \greencheck &  $22.8$ & $38.2$ \\ 

    \bottomrule
  \end{tabular}
  }
\label{tab:ablation}
\end{minipage}
\hspace{0.23cm}
\begin{minipage}[t]{0.38\linewidth}
\centering
\renewcommand\arraystretch{1.1}
\caption{ Effect of the voxel size $r$ and latent code size $l_{B}$ in detection and reconstruction on ScanNet-MDR dataset (mAP/IoU), and inference time per image.}
\resizebox{1\linewidth}{!}{
\begin{tabular}{|c|c|c|c||c|}\hline
\backslashbox{$r$}{$l_{B}$}   & $16$ & $32$ & $64$\tikzmark{end1}  &  \tabincell{c}{Time \\ ($ms$)}\\ \hline
 $0.64m$ & \tabincell{c}{$17.5/$\\ $30.7$} &
 \tabincell{c}{$17.7/$\\ $31.1$} &
 \tabincell{c}{$18.1/$\\ $31.4$} & $44.2$\\ \hline
 $0.32m$ & \tabincell{c}{$19.1/$\\ $34.1$} &
 \tabincell{c}{$18.8/$\\ $34.8$} &
 \tabincell{c}{$18.6/$\\ $35.3$} & $55.6$\\ \hline
 $0.16m$ \tikzmark{end} & \tabincell{c}{$21.9/$\\ $37.1$} &
 \tabincell{c}{$22.2/$\\ $37.6$} &
 \tabincell{c}{$\textbf{22.8}/$\\ $\textbf{38.2}$} & $78.1$
 \\ \hline

\end{tabular}
  
}

\label{tab:ablation_size}
\end{minipage}
\vspace{0mm}
\end{table}

\SubSection{Multiple Object Detection and Reconstruction on ScanNet-MDR}

\paragraph{Dataset.}
Since there is no benchmark providing both $3$D CAD models and $3$D bounding boxes for multiple objects within a single real image,  
we assemble a dataset with $18,000$ real images from the ScanNet~\cite{dai2017scannet}, termed ScanNet Monocular Detection and Reconstruction (ScanNet-MDR) dataset. 
For each object in an image, its CAD model is produced by~\cite{avetisyan2019scan2cad}. We then generate the corresponding $3$D bounding box label and camera calibration matrix.
Unlike the ShapeNet-pairs or ShapeNet-triplets datasets, all the $3$D objects in ScanNet-MDR are at absolute scale. Additionally, this dataset contains greater diversity including $19$ object categories (bag, basket, bathtub, bed, bench, bookshelf, cabinet, chair, display, file, lamp, microwave, piano, printer, sofa, stove, table, trash and washer). 
We split the data into $80\%$ for training and $20\%$ for testing.

\Paragraph{Experimental Settings.}
In this dataset, we use a voxel grid of size $10.28\times 3.2\times 6.4m$ ($
X=64, Y=20, Z=40$, $r=0.16m$), a minimum to encompass all annotated $3$D objects in the dataset. 
The PCA-SDF is pre-computed with $N_{S}\approx 560,000$ occupied voxels from $200$ training surfaces. 
We set $k=17$, $C=19$ and $l_{B}=64$. 
For comparison, we train CoReNet~\cite{popov2020corenet} using the released code on our training data. 
%
We use mAP with $3$D box IoU threshold of $0.15$ as detection metric~\cite{nie2020total3dunderstanding}, and global $3$D IoU on a $128^3$ voxel grid as reconstruction metric.

\Paragraph{Results and Ablation Studies.} 
We report detection and reconstruction results on the testing set. 
As shown in Tab.~\ref{tab:ablation}, our method significantly improves over CoReNet~\cite{popov2020corenet} on $3$D reconstruction and advances the ablated versions on both detection and reconstruction. 
Qualitative results are shown in Fig.~\ref{fig:scannet_reconstruction}. 
CoReNet, as an image-to-voxel reconstruction network without special design for feature smearing issue, cannot handle truncated objects in the input image. 


\emph{Joint Framework $vs.$ Separate Modules.}
Moreover, Tab.~\ref{tab:ablation} shows the ablation results of our models without detection (Proposed-$1$) or reconstruction (Proposed-$2$) modules, where one can conclude that joint framework in this work performs better than solving either task exclusively.

\emph{CenterNet-$3$D and PCA-SDF.}
To further validate the effectiveness of the proposed CenterNet-$3$D detector over the conventional CenterNet, we train a model (Proposed-$3$) by combining the reconstruction module with the conventional CenterNet, which formulates $3$D detection as a problem of $2$D keypoint detection  directly from the pixel-based image features. As compared in Tab.~\ref{tab:ablation}, our model outperforms Proposed-$3$ in both detection and reconstruction.
To compare PCA-SDF with DeepLS~\cite{chabra2020deep} in the joint detection and reconstruction framework, we train a model (Proposed-$4$) by using DeepLS representation as our fine-level reconstruction module.  Tab.~\ref{tab:ablation} shows that both detection and reconstruction performances are worse than ours.

\emph{Effect on Voxel Size $r$ Latent Code Size $l_{B}$.}
The validity of local shape pair, expressed by PCA-SDF in our work, depends on the voxel size. For instance, we show the percentage of explained variation for three voxel sizes in Fig.~\ref{fig:pca}. 
As larger voxel sizes are used, the first few bases could explain less variation, due to the diminished local shape similarity among larger voxels, {\it i.e.}, weakened local shape prior.
This is also validated by the ablation of voxel size $r$ and latent code size $l_B$ in Tab.~\ref{tab:ablation_size}, where larger voxel sizes lead to lower detection and reconstruction accuracies. 
On the other hand, a larger latent code size results in better representation power (Fig.~\ref{fig:pca}) but not necessarily reconstruction, since it imposes a more challenging  task for the network to predict a higher-dim code.

\emph{Effect on Positional Encoding.} To investigate the effect of positional encoding operator on $3$D detection and reconstruction, we retrain a model without the positional encoding (Proposed-w/o PE). As compared in Tab.~\ref{tab:ablation}, both detection and reconstruction accuracies are worse than ours, which indicates that the positional encoding indeed enhances the voxel feature representation.

\emph{Computation Time.} 
Tab.~\ref{tab:ablation_size} validates our inference time per image with different voxel sizes on a GTX 1080Ti GPU. Since $l_{B}$ does not affect the runtime much, we show the average time across  three $l_{B}$.

\SubSection{3D Shape Representation Power of PCA-SDF}

While PCA-SDF has demonstrated its advantage in $2$D to $3$D reconstruction, this is rooted from its ability in representing $3$D shapes.
To quantify its $3$D shape representation power, we design the following experiment and compare with DeepSDF~\cite{park2019deepsdf} and DeepLS~\cite{chabra2020deep}, without involving $2$D image inputs.
Following the setting of~\cite{chabra2020deep}, we utilize $1,000$ ShapeNet shapes ($200$ each from $5$ categories) to compute our PCA-SDF bases. 
Each $3$D shape is split by a $32{\times}32{\times}32$ grids ($r=\frac{1}{32}$). 
During training, both DeepSDF and DeepLS optimize the latent codes and decoder weights through backpropagation, to best represent the training shapes. 
In inference, decoder weights are fixed, and the optimal latent code is estimated given a testing shape. 
In contrast, 
we compute the latent codes whose multiplication with PCA-SDF bases can best approximate the ground-truth SDF.


We evaluate $3$D shape reconstruction accuracy on various categories. 
As shown in Tab.~\ref{tab:shape_representation}, PCA-SDF has lower reconstruction error than DeepLS, even with a smaller number of representation parameters. 
Moreover, our inference is $10\times$ more efficient (infer at $256^3$ resolution) which meets the real-time requirement for downstream tasks. 

To study whether the PCA-based representation is sensitive to tiny geometry perturbation, we apply minimal translation ($\pm0.1r$, $\pm0.05r$) and rotation ($\pm4^{\circ}$, $\pm2^{\circ}$) to testing surfaces of the $5$ categories and evaluate surface reconstruction error on these data, while no data augmentation was applied to the training data of shape bases computing. 
As shown in Fig.~\ref{fig:sensitive}, the error is stable in a very small range, which illustrates that PCA-SDF is robust to translation and rotation variations.

\emph{Generalization to Unseen Category.} 
In order to investigate the generalization of PCA-SDF, we design an experiment to compute PCA-SDF from a single category ($200$ shapes), and reconstruct $3$D shapes from the other four unseen categories. 
We repeat the training/testing $5$ times across $5$ categories.
As reported in Tab.~\ref{tab:shape_representation}, PCA-SDF trained on unseen categories achieves a comparable performance ($0.028$ {\it vs.}~$0.022$) with the one trained on seen categories when $l_{B}=125$, which indicates that local $3$D shapes at the voxel level are indeed similar to each other, even across different categories.


\begin{table}[t!]
 \renewcommand{\arraystretch}{1.2}
  \caption{ Comparison of reconstructing $3$D shapes from ShapeNet test set, evaluated by Chamfer Distance-$L2$ (multiplied by $10^{3}$).  PCA-SDF achieves higher accuracy and efficiency than DeepLS even with fewer decoder and representation parameters. Decoder para.~refer to the decoder network parameters for DeepSDF or DeepLS, and PCA bases for our PCA-SDF. [Key: \firstkey{Best}, \secondkey{Second Best}]}
  \centering
  \resizebox{1\linewidth}{!}{
  \begin{tabular}{l | c c c c c |c|| c || c| c| c}
    \toprule
    Method & Chair & Plane & Table & Lamp & Sofa & Mean &  Unseen & \tabincell{c}{\#Decoder\\Para. (M)}   &  \tabincell{c}{\#Represent. \\Para. (K)}  & \tabincell{c}{Inference \\Time (s)}   \\
    \midrule
    DeepSDF~\cite{park2019deepsdf}  & $0.204$  & $0.143$ & $0.553$ & $0.832$ & $0.132$ & $0.372$ & - & $1.8$ &  $\textbf{0.3}$ & $6.9626$      \\ 
    DeepLS~\cite{chabra2020deep} ($l_{B}{=}125$)
     & $0.030$ & $0.018$ & $0.032$ & $0.078$ & $0.044$ & $0.040$ &- & $0.05$ & $4096$ & $0.8081$      \\ \hline
    PCA-SDF ($l_{B}{=}32$)
     & $0.031$  & $0.016$ & $0.033$ & $0.035$ & $0.032$ & $0.029$  & $0.111$  & $\textbf{0.02}$ & $1049$ & $\textbf{0.0126}$      \\ 
    PCA-SDF ($l_{B}{=}64$) & $\secondkey{0.027}$  & $\secondkey{0.012}$ & $\secondkey{0.030}$ & $\secondkey{0.027}$ & $\secondkey{0.030}$ & $\secondkey{0.025}$ & $0.059$  & $0.05$ & $2097$ & $0.0129$     \\ 
    PCA-SDF ($l_{B}{=}125$)  
    & $\firstkey{0.026}$  & $\firstkey{0.010}$ & $\firstkey{0.029}$ & $\firstkey{0.016}$ & $\firstkey{0.029}$ & $\firstkey{0.022}$ & $\textbf{0.028}$ & $0.09$ & $4096$ & $0.0132$      \\ 
        \bottomrule
  \end{tabular}
  }
  \label{tab:shape_representation}
  \vspace{0mm}
\end{table}


\Section{Conclusion}
We present a voxel-based $3$D detection and reconstruction framework for predicting $3$D bounding boxes and shapes of multiple objects from a single image. Specifically, we first learn a regular grid of $3$D voxel features for the input images. 
Based on the voxel features, we devise a novel CenterNet-$3$D detector to detect and regress $3$D bounding boxes in the $3$D space.
With a coarse-level voxelization and a fine-level local PCA-SDF representation, our reconstruction module provides highly efficient and accurate reconstructions.  
The comprehensive experiments  show the superiority of the proposed method in $3$D detection and reconstruction, as well as shape representation power. 
The same as CoReNet and Points2Objects, one limitation of our approach is that it requires the camera calibration matrix as input which might limit its application to real images.  
Therefore, one future direction is to invest the necessity of this requirement and/or integrate with auto-calibration methods.
%


\bibliographystyle{unsrt}
\bibliography{refs}

\begin{thebibliography}{10}

\bibitem{guerry2017snapnet}
Joris Guerry, Alexandre Boulch, Bertrand Le~Saux, Julien Moras, Aur{\'e}lien
  Plyer, and David Filliat.
\newblock Snapnet-r: Consistent 3{D} multi-view semantic labeling for robotics.
\newblock In {\em ICCV}, 2017.

\bibitem{tateno2017cnn}
Keisuke Tateno, Federico Tombari, Iro Laina, and Nassir Navab.
\newblock {CNN-SLAM}: Real-time dense monocular slam with learned depth
  prediction.
\newblock In {\em CVPR}, 2017.

\bibitem{han2020live}
Lei Han, Tian Zheng, Yinheng Zhu, Lan Xu, and Lu~Fang.
\newblock Live semantic 3{D} perception for immersive augmented reality.
\newblock {\em TVCG}, 2020.

\bibitem{behl2017bounding}
Aseem Behl, Omid Hosseini~Jafari, Siva Karthik~Mustikovela, Hassan Abu~Alhaija,
  Carsten Rother, and Andreas Geiger.
\newblock Bounding boxes, segmentations and object coordinates: How important
  is recognition for 3{D} scene flow estimation in autonomous driving
  scenarios?
\newblock In {\em ICCV}, 2017.

\bibitem{chen2018lidar}
Yiping Chen, Jingkang Wang, Jonathan Li, Cewu Lu, Zhipeng Luo, Han Xue, and
  Cheng Wang.
\newblock {LiDAR}-video driving dataset: Learning driving policies effectively.
\newblock In {\em CVPR}, 2018.

\bibitem{geiger2012we}
Andreas Geiger, Philip Lenz, and Raquel Urtasun.
\newblock Are we ready for autonomous driving? the {KITTI} vision benchmark
  suite.
\newblock In {\em CVPR}, 2012.

\bibitem{chen2016monocular}
Xiaozhi Chen, Kaustav Kundu, Ziyu Zhang, Huimin Ma, Sanja Fidler, and Raquel
  Urtasun.
\newblock Monocular 3{D} object detection for autonomous driving.
\newblock In {\em CVPR}, 2016.

\bibitem{huang2018cooperative}
Siyuan Huang, Siyuan Qi, Yinxue Xiao, Yixin Zhu, Ying~Nian Wu, and Song-Chun
  Zhu.
\newblock Cooperative holistic scene understanding: Unifying 3{D} object,
  layout, and camera pose estimation.
\newblock In {\em NeurIPS}, 2018.

\bibitem{brazil2019m3d}
Garrick Brazil and Xiaoming Liu.
\newblock {M3D-RPN}: Monocular 3{D} region proposal network for object
  detection.
\newblock In {\em ICCV}, 2019.

\bibitem{chen2020monopair}
Yongjian Chen, Lei Tai, Kai Sun, and Mingyang Li.
\newblock Monopair: Monocular 3{D} object detection using pairwise spatial
  relationships.
\newblock In {\em CVPR}, 2020.

\bibitem{kinematic-3d-object-detection-in-monocular-video}
Garrick Brazil, Gerard Pons-Moll, Xiaoming Liu, and Bernt Schiele.
\newblock Kinematic 3d object detection in monocular video.
\newblock In {\em CVPR}, 2020.

\bibitem{groomed-nms-grouped-mathematically-differentiable-nms-for-monocular-3d-object-detection}
Abhinav Kumar, Garrick Brazil, and Xiaoming Liu.
\newblock Groomed-nms: Grouped mathematically differentiable nms for monocular
  3{D} object detection.
\newblock In {\em CVPR}, 2021.

\bibitem{wu2017marrnet}
Jiajun Wu, Yifan Wang, Tianfan Xue, Xingyuan Sun, Bill Freeman, and Josh
  Tenenbaum.
\newblock Marrnet: 3{D} shape reconstruction via 2.5{D} sketches.
\newblock In {\em NeurIPS}, 2017.

\bibitem{zhu2018visual}
Jun-Yan Zhu, Zhoutong Zhang, Chengkai Zhang, Jiajun Wu, Antonio Torralba, Josh
  Tenenbaum, and Bill Freeman.
\newblock Visual object networks: Image generation with disentangled 3{D}
  representations.
\newblock In {\em NeurIPS}, 2018.

\bibitem{chen2018learning}
Zhiqin Chen and Hao Zhang.
\newblock Learning implicit fields for generative shape modeling.
\newblock In {\em CVPR}, 2019.

\bibitem{groueix2018atlasnet}
Thibault Groueix, Matthew Fisher, Vladimir~G Kim, Bryan~C Russell, and Mathieu
  Aubry.
\newblock Atlasnet: A papier-m{\^a}ch{\'e} approach to learning 3{D} surface
  generation.
\newblock In {\em CVPR}, 2018.

\bibitem{wen2019pixel2mesh++}
Chao Wen, Yinda Zhang, Zhuwen Li, and Yanwei Fu.
\newblock {Pixel2Mesh++:} multi-view 3{D} mesh generation via deformation.
\newblock In {\em ICCV}, 2019.

\bibitem{kundu20183d}
Abhijit Kundu, Yin Li, and James~M Rehg.
\newblock {3D-RCNN}: Instance-level 3{D} object reconstruction via
  render-and-compare.
\newblock In {\em CVPR}, 2018.

\bibitem{nie2020total3dunderstanding}
Yinyu Nie, Xiaoguang Han, Shihui Guo, Yujian Zheng, Jian Chang, and Jian~Jun
  Zhang.
\newblock Total3{DU}nderstanding: Joint layout, object pose and mesh
  reconstruction for indoor scenes from a single image.
\newblock In {\em CVPR}, 2020.

\bibitem{runz2020frodo}
Martin Runz, Kejie Li, Meng Tang, Lingni Ma, Chen Kong, Tanner Schmidt, Ian
  Reid, Lourdes Agapito, Julian Straub, Steven Lovegrove, et~al.
\newblock {FroDO}: From detections to 3{D} objects.
\newblock In {\em CVPR}, 2020.

\bibitem{meshrcnn}
Justin~Johnson Georgia~Gkioxari, Jitendra~Malik.
\newblock Mesh {R-CNN}.
\newblock In {\em ICCV}, 2019.

\bibitem{popov2020corenet}
Stefan Popov, Pablo Bauszat, and Vittorio Ferrari.
\newblock {CoReNet}: Coherent 3{D} scene reconstruction from a single {RGB}
  image.
\newblock In {\em ECCV}, 2020.

\bibitem{engelmann2020points}
Francis Engelmann, Konstantinos Rematas, Bastian Leibe, and Vittorio Ferrari.
\newblock From points to multi-object 3{D} reconstruction.
\newblock In {\em CVPR}, 2021.

\bibitem{sitzmann2019deepvoxels}
Vincent Sitzmann, Justus Thies, Felix Heide, Matthias Nie{\ss}ner, Gordon
  Wetzstein, and Michael Zollhofer.
\newblock {DeepVoxels}: Learning persistent 3{D} feature embeddings.
\newblock In {\em CVPR}, 2019.

\bibitem{roddick2018orthographic}
Thomas Roddick, Alex Kendall, and Roberto Cipolla.
\newblock Orthographic feature transform for monocular 3{D} object detection.
\newblock In {\em BMVC}, 2018.

\bibitem{reading2021categorical}
Cody Reading, Ali Harakeh, Julia Chae, and Steven~L Waslander.
\newblock Categorical depth distribution network for monocular 3{D} object
  detection.
\newblock In {\em CVPR}, 2021.

\bibitem{guillard2020uclid}
Benoit Guillard, Edoardo Remelli, and Pascal Fua.
\newblock {UCLID-N}et: Single view reconstruction in object space.
\newblock In {\em NeurIPS}, 2020.

\bibitem{zhou2019objects}
Xingyi Zhou, Dequan Wang, and Philipp Kr{\"a}henb{\"u}hl.
\newblock Objects as points.
\newblock {\em arXiv preprint arXiv:1904.07850}, 2019.

\bibitem{engelmann2016joint}
Francis Engelmann, J{\"o}rg St{\"u}ckler, and Bastian Leibe.
\newblock Joint object pose estimation and shape reconstruction in urban street
  scenes using 3{D} shape priors.
\newblock In {\em GCPR}, 2016.

\bibitem{chabra2020deep}
Rohan Chabra, Jan~E Lenssen, Eddy Ilg, Tanner Schmidt, Julian Straub, Steven
  Lovegrove, and Richard Newcombe.
\newblock Deep local shapes: Learning local {SDF} priors for detailed 3{D}
  reconstruction.
\newblock In {\em ECCV}, 2020.

\bibitem{zeeshan2014cars}
Muhammad Zeeshan~Zia, Michael Stark, and Konrad Schindler.
\newblock Are cars just 3{D} boxes?-jointly estimating the 3d shape of multiple
  objects.
\newblock In {\em CVPR}, 2014.

\bibitem{chabot2017deep}
Florian Chabot, Mohamed Chaouch, Jaonary Rabarisoa, C{\'e}line Teuliere, and
  Thierry Chateau.
\newblock Deep {MANTA}: A coarse-to-fine many-task network for joint 2{D} and
  3{D} vehicle analysis from monocular image.
\newblock In {\em CVPR}, 2017.

\bibitem{li2017deep}
Chi Li, M~Zeeshan~Zia, Quoc-Huy Tran, Xiang Yu, Gregory~D Hager, and Manmohan
  Chandraker.
\newblock Deep supervision with shape concepts for occlusion-aware 3{D} object
  parsing.
\newblock In {\em CVPR}, 2017.

\bibitem{su2015render}
Hao Su, Charles~R Qi, Yangyan Li, and Leonidas~J Guibas.
\newblock Render for {CNN}: Viewpoint estimation in images using cnns trained
  with rendered 3{D} model views.
\newblock In {\em ICCV}, 2015.

\bibitem{tulsiani2015viewpoints}
Shubham Tulsiani and Jitendra Malik.
\newblock Viewpoints and keypoints.
\newblock In {\em CVPR}, 2015.

\bibitem{xiang2015data}
Yu~Xiang, Wongun Choi, Yuanqing Lin, and Silvio Savarese.
\newblock Data-driven 3{D} voxel patterns for object category recognition.
\newblock In {\em CVPR}, 2015.

\bibitem{zia2015towards}
M~Zeeshan Zia, Michael Stark, and Konrad Schindler.
\newblock Towards scene understanding with detailed 3{D} object
  representations.
\newblock {\em IJCV}, 112(2):188--203, 2015.

\bibitem{poirson2016fast}
Patrick Poirson, Phil Ammirato, Cheng-Yang Fu, Wei Liu, Jana Kosecka, and
  Alexander~C Berg.
\newblock Fast single shot detection and pose estimation.
\newblock In {\em 3DV}, 2016.

\bibitem{atoum2017monocular}
Yousef Atoum, Joseph Roth, Michael Bliss, Wende Zhang, and Xiaoming Liu.
\newblock Monocular video-based trailer coupler detection using multiplexer
  convolutional neural network.
\newblock In {\em ICCV}, 2017.

\bibitem{chen20153d}
Xiaozhi Chen, Kaustav Kundu, Yukun Zhu, Andrew~G Berneshawi, Huimin Ma, Sanja
  Fidler, and Raquel Urtasun.
\newblock 3{D} object proposals for accurate object class detection.
\newblock In {\em NeurIPS}, 2015.

\bibitem{xu2018multi}
Bin Xu and Zhenzhong Chen.
\newblock Multi-level fusion based 3{D} object detection from monocular images.
\newblock In {\em CVPR}, 2018.

\bibitem{simonelli2019disentangling}
Andrea Simonelli, Samuel~Rota Bulo, Lorenzo Porzi, Manuel L{\'o}pez-Antequera,
  and Peter Kontschieder.
\newblock Disentangling monocular 3{D} object detection.
\newblock In {\em ICCV}, 2019.

\bibitem{liu2019deep}
Lijie Liu, Jiwen Lu, Chunjing Xu, Qi~Tian, and Jie Zhou.
\newblock Deep fitting degree scoring network for monocular 3{D} object
  detection.
\newblock In {\em CVPR}, 2019.

\bibitem{choi2013understanding}
Wongun Choi, Yu-Wei Chao, Caroline Pantofaru, and Silvio Savarese.
\newblock Understanding indoor scenes using 3{D} geometric phrases.
\newblock In {\em CVPR}, 2013.

\bibitem{huang2018holistic}
Siyuan Huang, Siyuan Qi, Yixin Zhu, Yinxue Xiao, Yuanlu Xu, and Song-Chun Zhu.
\newblock Holistic 3{D} scene parsing and reconstruction from a single {RGB}
  image.
\newblock In {\em ECCV}, 2018.

\bibitem{chang2015shapenet}
Angel~X. Chang, Thomas Funkhouser, Leonidas Guibas, Pat Hanrahan, Qixing Huang,
  Zimo Li, Silvio Savarese, Manolis Savva, Shuran Song, Hao Su, Jianxiong Xiao,
  Li~Yi, and Fisher Yu.
\newblock Shape{N}et: An information-rich 3{D} model repository.
\newblock {\em arXiv preprint arXiv:1512.03012}, 2015.

\bibitem{jiang2019disentangled}
Zi-Hang Jiang, Qianyi Wu, Keyu Chen, and Juyong Zhang.
\newblock Disentangled representation learning for 3{D} face shape.
\newblock In {\em CVPR}, 2019.

\bibitem{ranjan2018generating}
Anurag Ranjan, Timo Bolkart, Soubhik Sanyal, and Michael~J Black.
\newblock Generating 3{D} faces using convolutional mesh autoencoders.
\newblock In {\em ECCV}, 2018.

\bibitem{liu20193d}
Feng Liu, Luan Tran, and Xiaoming Liu.
\newblock 3{D} face modeling from diverse raw scan data.
\newblock In {\em ICCV}, 2019.

\bibitem{bagautdinov2018modeling}
Timur Bagautdinov, Chenglei Wu, Jason Saragih, Pascal Fua, and Yaser Sheikh.
\newblock Modeling facial geometry using compositional {VAE}s.
\newblock In {\em CVPR}, 2018.

\bibitem{dai2017shape}
Angela Dai, Charles Ruizhongtai~Qi, and Matthias Nie{\ss}ner.
\newblock Shape completion using 3{D}-encoder-predictor {CNN}s and shape
  synthesis.
\newblock In {\em CVPR}, 2017.

\bibitem{stutz2018learning}
David Stutz and Andreas Geiger.
\newblock Learning 3{D} shape completion from laser scan data with weak
  supervision.
\newblock In {\em CVPR}, 2018.

\bibitem{fully-understanding-generic-objects-modeling-segmentation-and-reconstruction}
Feng Liu, Luan Tran, and Xiaoming Liu.
\newblock Fully understanding generic objects: Modeling, segmentation, and
  reconstruction.
\newblock In {\em CVPR}, 2021.

\bibitem{he2017mask}
Kaiming He, Georgia Gkioxari, Piotr Doll{\'a}r, and Ross Girshick.
\newblock Mask {R-CNN}.
\newblock In {\em ICCV}, 2017.

\bibitem{qi2017pointnet}
Charles~R Qi, Hao Su, Kaichun Mo, and Leonidas~J Guibas.
\newblock Pointnet: Deep learning on point sets for 3{D} classification and
  segmentation.
\newblock In {\em CVPR}, 2017.

\bibitem{wang2018pixel2mesh}
Nanyang Wang, Yinda Zhang, Zhuwen Li, Yanwei Fu, Wei Liu, and Yu-Gang Jiang.
\newblock Pixel2mesh: Generating 3{D} mesh models from single {RGB} images.
\newblock In {\em ECCV}, 2018.

\bibitem{choy20163d}
Christopher~B Choy, Danfei Xu, JunYoung Gwak, Kevin Chen, and Silvio Savarese.
\newblock 3{D}-{R}2{N}2: A unified approach for single and multi-view 3{D}
  object reconstruction.
\newblock In {\em ECCV}, 2016.

\bibitem{mescheder2018occupancy}
Lars Mescheder, Michael Oechsle, Michael Niemeyer, Sebastian Nowozin, and
  Andreas Geiger.
\newblock Occupancy networks: Learning 3{D} reconstruction in function space.
\newblock In {\em CVPR}, 2019.

\bibitem{park2019deepsdf}
Jeong~Joon Park, Peter Florence, Julian Straub, Richard Newcombe, and Steven
  Lovegrove.
\newblock Deep{SDF}: Learning continuous signed distance functions for shape
  representation.
\newblock In {\em CVPR}, 2019.

\bibitem{learning-implicit-functions-for-topology-varying-dense-3d-shape-correspondence}
Feng Liu and Xiaoming Liu.
\newblock Learning implicit functions for topology-varying dense 3{D} shape
  correspondence.
\newblock In {\em NeurIPS}, 2020.

\bibitem{genova2019learning}
Kyle Genova, Forrester Cole, Daniel Vlasic, Aaron Sarna, William~T Freeman, and
  Thomas Funkhouser.
\newblock Learning shape templates with structured implicit functions.
\newblock In {\em ICCV}, 2019.

\bibitem{deng2020cvxnet}
Boyang Deng, Kyle Genova, Soroosh Yazdani, Sofien Bouaziz, Geoffrey Hinton, and
  Andrea Tagliasacchi.
\newblock {CvxNet}: Learnable convex decomposition.
\newblock In {\em CVPR}, 2020.

\bibitem{chen2020bsp}
Zhiqin Chen, Andrea Tagliasacchi, and Hao Zhang.
\newblock {BSP-Net}: Generating compact meshes via binary space partitioning.
\newblock In {\em CVPR}, 2020.

\bibitem{jiang2020local}
Chiyu Jiang, Avneesh Sud, Ameesh Makadia, Jingwei Huang, Matthias Nie{\ss}ner,
  Thomas Funkhouser, et~al.
\newblock Local implicit grid representations for 3{D} scenes.
\newblock In {\em CVPR}, 2020.

\bibitem{takikawa2021neural}
Towaki Takikawa, Joey Litalien, Kangxue Yin, Karsten Kreis, Charles Loop, Derek
  Nowrouzezahrai, Alec Jacobson, Morgan McGuire, and Sanja Fidler.
\newblock Neural geometric level of detail: Real-time rendering with implicit
  3{D} shapes.
\newblock In {\em CVPR}, 2021.

\bibitem{liu2020neural}
Lingjie Liu, Jiatao Gu, Kyaw~Zaw Lin, Tat-Seng Chua, and Christian Theobalt.
\newblock Neural sparse voxel fields.
\newblock In {\em NeurIPS}, 2020.

\bibitem{michalkiewicz2020simple}
Mateusz Michalkiewicz, Eugene Belilovsky, Mahsa Baktashmotlagh, and Anders
  Eriksson.
\newblock A simple and scalable shape representation for 3d reconstruction.
\newblock In {\em BMVC}, 2020.

\bibitem{ricao2017compressed}
Daniel Ricao~Canelhas, Erik Schaffernicht, Todor Stoyanov, Achim~J Lilienthal,
  and Andrew~J Davison.
\newblock Compressed voxel-based mapping using unsupervised learning.
\newblock {\em Robotics}, 2017.

\bibitem{tang2018real}
Danhang Tang, Mingsong Dou, Peter Lincoln, Philip Davidson, Kaiwen Guo,
  Jonathan Taylor, Sean Fanello, Cem Keskin, Adarsh Kowdle, Sofien Bouaziz,
  et~al.
\newblock Real-time compression and streaming of 4{D} performances.
\newblock {\em TOG}, 2018.

\bibitem{vaswani2017attention}
Ashish Vaswani, Noam Shazeer, Niki Parmar, Jakob Uszkoreit, Llion Jones,
  Aidan~N Gomez, {\L}ukasz Kaiser, and Illia Polosukhin.
\newblock Attention is all you need.
\newblock In {\em NeurIPS}, 2017.

\bibitem{cciccek20163d}
{\"O}zg{\"u}n {\c{C}}i{\c{c}}ek, Ahmed Abdulkadir, Soeren~S Lienkamp, Thomas
  Brox, and Olaf Ronneberger.
\newblock {3D U-Net}: learning dense volumetric segmentation from sparse
  annotation.
\newblock In {\em MICCAI}, 2016.

\bibitem{liu2020smoke}
Zechen Liu, Zizhang Wu, and Roland T{\'o}th.
\newblock Smoke: Single-stage monocular 3d object detection via keypoint
  estimation.
\newblock In {\em CVPRW}, 2020.

\bibitem{lin2017focal}
Tsung-Yi Lin, Priya Goyal, Ross Girshick, Kaiming He, and Piotr Doll{\'a}r.
\newblock Focal loss for dense object detection.
\newblock In {\em ICCV}, 2017.

\bibitem{sun2018pix3d}
Xingyuan Sun, Jiajun Wu, Xiuming Zhang, Zhoutong Zhang, Chengkai Zhang, Tianfan
  Xue, Joshua~B Tenenbaum, and William~T Freeman.
\newblock Pix3{D}: Dataset and methods for single-image 3{D} shape modeling.
\newblock In {\em CVPR}, 2018.

\bibitem{dai2017scannet}
Angela Dai, Angel~X Chang, Manolis Savva, Maciej Halber, Thomas Funkhouser, and
  Matthias Nie{\ss}ner.
\newblock Scannet: Richly-annotated 3{D} reconstructions of indoor scenes.
\newblock In {\em CVPR}, 2017.

\bibitem{avetisyan2019scan2cad}
Armen Avetisyan, Manuel Dahnert, Angela Dai, Manolis Savva, Angel~X Chang, and
  Matthias Nie{\ss}ner.
\newblock {Scan2CAD}: Learning {CAD} model alignment in {RGB-D} scans.
\newblock In {\em CVPR}, 2019.

\end{thebibliography}


\end{document}